\documentclass[letterpaper, 10 pt, conference]{ieeeconf}
\IEEEoverridecommandlockouts
\usepackage{amsmath,amssymb,graphicx, subfigure,epstopdf,cite,enumerate,booktabs,algorithm,algorithmic,setspace,float}

\title{Decentralized Cooperative Multi-Robot Localization with EKF}
{\vspace{6mm}}
\author{Ruihua Han, Shengduo Chen, Yasheng Bu, Zhijun Lyu and Qi Hao*
\thanks{*This work was supported by the Science and Technology Innovation Commission of Shenzhen Municipality (No. CKFW2016041415372174 and No. GJHZ20170314114424152), and the National Natural Science Foundation of China (No. 61773197).
}
\thanks{The authors are with Department of Computer Science and Engineering, Southern University of Science and Technology, Shenzhen, Guangdong, 518055, China.
        \{{\tt\small hanrh@mail.sustc.edu.cn}
        {\tt\small haoq@sustc.edu.cn}\} *Corresponding author}%
}

\begin{document}
\maketitle
\thispagestyle{empty}
\pagestyle{empty}

\begin{abstract}
Multi-robot localization has been a critical problem for robots performing complex tasks cooperatively. In this paper, we propose a decentralized approach to localize a group of robots in a large featureless environment. The proposed approach only requires that at least one robot remains stationary as a temporary landmark during a certain period of time. The novelty of our approach is threefold: (1) developing a decentralized scheme that each robot calculates their own state and only stores the latest one to reduce storage and computational cost; (2) developing an efficient localization algorithm through the extended Kalman filter (EKF) that only uses observations of relative pose to estimate the robot positions; (3) developing a scheme has less requirements on landmarks and more robustness against insufficient observations. Various simulations and experiments using five robots equipped with relative pose-measurement sensors are performed to validate the superior performance of our approach.
\end{abstract}


\IEEEpeerreviewmaketitle

\section{Introduction}

In the study of autonomous mobile robots, localization is one of the most fundamental problems\cite{Cox1991}. Mobile robot localization, as known as position estimation, is the process of determining the pose of a robot relative to a given map of the environment. Compared to a single robot, multi-robot systems are more robust and efficient to perform complex tasks within complex environments, including rescue and disaster management, surveillance and monitoring, and so on\cite{Saeedi2016}. For the autonomous multi-robot systems, the performance of most functionalities heavily relies on the accuracy of robot localization results, where individual localization results easily interfere with each other.

The main multi-robot localization approaches can fall into three categories: (1) external, (2) internal, and (3) cooperative. The external methods help each robot to acquire their position information independently through those external active landmarks such as Global Position Systems (GPS) or Ultra-Wideband (UWB) systems. The localization uncertainty of the external methods can be bounded only when those external signals are reliable and available. The internal methods enable individual robots to localize their positions by utilizing the sensors on each robot such as camera or laser scanner to detect the features of the environment. However, these methods can not work well in those environments without rich feature information. The cooperative methods, also known as cooperative localization (CL), allow a group of communicating robots to estimate their states jointly using relative observations of each other. Those methods are advantageous in their low cost and high scalability, receiving increasing attentions recently, especially for those robot systems working in featureless environments.

\begin{figure}
\centering
        \includegraphics[width=0.45\textwidth,clip]{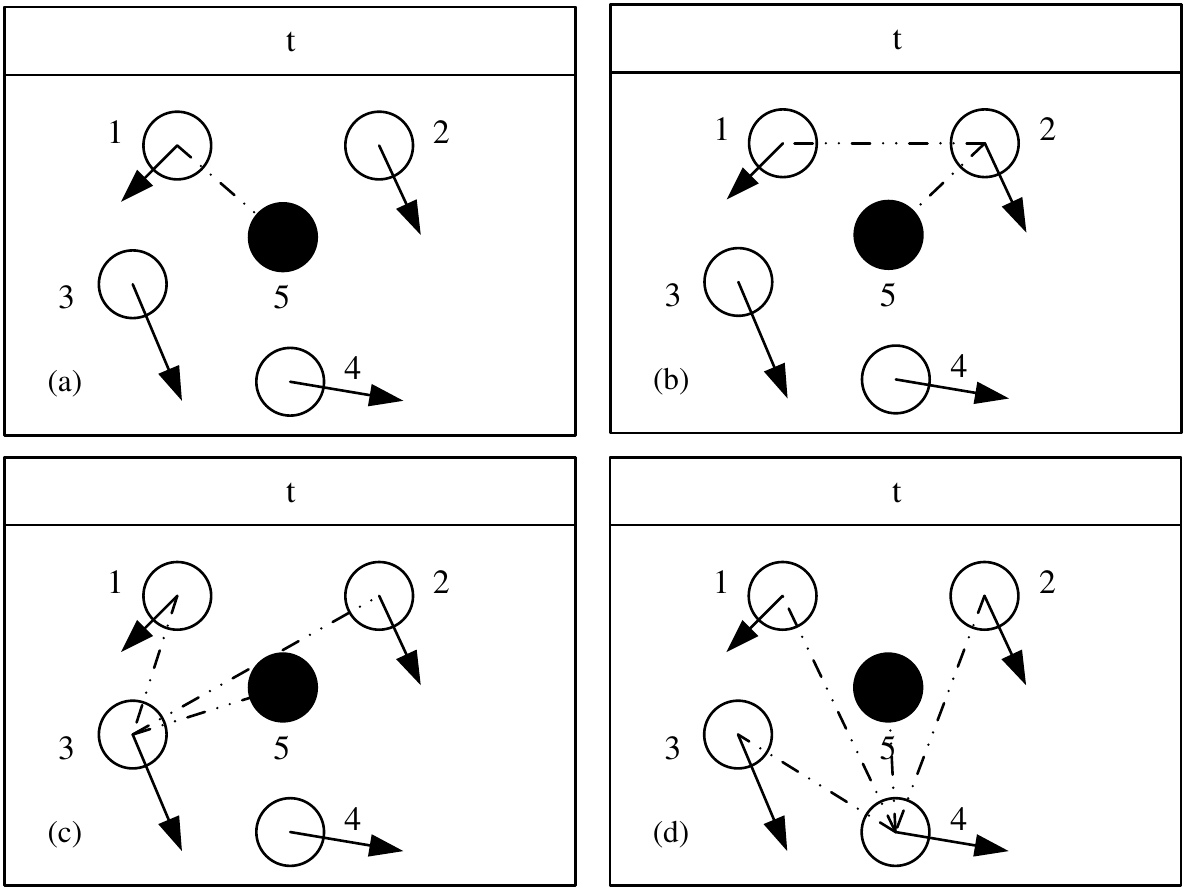}
    \caption{Illustration of the approach. The example during (a) denotes at the time t, one robot is randomly chose to remain stationary, firstly, the nearest moving robot 1 detect the stationary one to estimate its pose. During (b), the next nearest robot to robot 1 estimate its pose by the two observation. During (c) and (d) the robot estimate their pose by the observation of previous robot.}
    \label{system}
\end{figure}

However, many technical challenges need to be overcome in using cooperative localization methods. Generally speaking, the estimation accuracy in featureless environments is inferior to that in environments with rich feature information. Besides, as the group of robots becomes larger, the cost of computation and communication grows much higher. In practice, only using relative observations of each other may not be enough to localize multiple robots given various uncertainties and noises.

A number of approaches have been proposed to address above challenges. Dead-reckoning(DR) is one common method which estimates the robot pose from its previous pose information by using proprioceptive sensors\cite{borenstein1994umbmark}, but the estimation errors will be accumulated over a long distance. To improve the accuracy and consistency of estimated robot states, various data fusion algorithms have been developed, such as Markov localization (ML), extended Kalman filter (EKF) localization and Monte Carlo localization (MCL)\cite{Everett1994}. The main idea of these schemes is to optimally combine the data derived from proprioceptive and exteroceptive sensors and filter out the motion and measurement noises to reduce the uncertainty of robot pose estimation. Usually, the dependency on landmarks limits the applications of those approaches.


Generally, the schemes to calculate the data fusion in CL are either centralized, multi-centralized or decentralized\cite{Masinjila2018}. In a centralized system, the computations for all robot data are processed in one central robot. Central robot provides feedback and required sensory information from all other robots \cite{Howard2002}. The major disadvantages of this scheme include the high expense of computation and the high vulnerability of the whole system.  In a multi-centralized system, there are more than one robot performing as the central robot to process all the data of entire group\cite{Nerurkar2010}. Consequently, this system needs higher computational power. The decentralized system means each robot estimates its own pose respectively through processing their own sensors data and updates it locally. Nonetheless the problem of double-counting or data incest may occur when two uncorrelated robots share information\cite{Luft2018}. The decentralized system yield the least computational cost and highest system scalability.

In this paper, we present a decentralized approach for CL, the approach is illustrated as Fig. \ref{system}. In this approach, at each time instant only one random robot in the group remains stationary, while the other moving robots estimate their state by fusing the data of odometry and observations of relative observation between two robots. The rearward moving robots utilize the observation with the forward moving robots to improve the accuracy and stability of the estimator. And the order depends on the relative distance. The contributions of this work include:

1) A decentralized scheme for CL is proposed, where each robot estimates their own state that reduces the cost of computation.

2) An EKF Based algorithm is proposed, where each robot only maintains its own pose and corresponding covariance, and the communication between two robots only occurs when one robot detect another at each time instant, yielding a computational complexity of \emph{O}(\emph{N}).

3) This approach only requires the odometry data and relative pose measurements so that it can perform in GPS-denied featureless environments. Even though there are insufficient relative pose measurements, the proposed approach can still guarantee a moderate performance.

The rest of this paper is organized as follows. Section \uppercase\expandafter{\romannumeral2} describes the related work. Section \uppercase\expandafter{\romannumeral3} presents the system setup and the localization problem. Section \uppercase\expandafter{\romannumeral4} gives the details of the proposed algorithm. Section \uppercase\expandafter{\romannumeral5} provides simulation and experiment results and discussions. Section \uppercase\expandafter{\romannumeral6} concludes the paper and outlines future works.

\section{Related Work}

In the problem of multi-robot localization, the study of CL receives considerable attention for a long period, because of its scalability and superior performance. Many approaches on CL have been proposed based on various schemes and algorithms as mentioned above. In this section, we describe several relevant approaches on CL to highlight the novelty of this work.

To estimate the state of robots with high accuracy and consistency, various filter algorithms have been tested for CL. An EKF-based algorithm for heterogeneous outdoor multi-robot localization is described in \cite{madhavan2002distributed}, where individual robot equipped with encoders and a camera maintains an estimated pose by using sensor data fusion through EKF.  An improved EKF on CL is studied in \cite{Martinelli2005}. Additionally, a recent approach that called recursive decentralized localization based on EKF is presented in \cite{Luft2018}. The proposed algorithm approximates the inter-robot correlations and performs with asynchronous pairwise communication. Other algorithms such as particle filter (PF) \cite{howard2006multi} and maximum a posteriori (MAP)\cite{Nerurkar2009} are also extensively studied for CL. The main limitation of these approaches is that in certain case, none of the robots in group has an access to the absolute state information that will reduce the estimation accuracy and consistency. In our approach, the stationary robot at each time can improve the accuracy efficiently by an optimized EKF algorithm.

To reduce the cost of computation, various decentralized schemes are proposed. Luis C \emph{et al.} present a Covariance Intersection (CI)-based algorithm for reducing the complexity of CL\cite{Franchi2013}, where each robot locally maintains its state and covariance. Amanda \emph{et al.} provide a PF-based algorithm that is completely decentralized and a low-cost particle clustering method to reduce the computational complexity  \cite{prorok2012low}. In a decentralized system, it is important to avoid the problem of double-counting or data incest. Masinjila \emph{et al.} presents a heuristically tuned EKF to address this problem \cite{Masinjila2018}, which proposes an empirical methodology for improving the consistency of EKF estimates by means of artificial inflation of the landmark covariance. This problem is also discussed in \cite{luft2016recursive} and \cite{Luft2018}. Our approach also employs the similar decentralized scheme to reduce the system complexity.

The approach of CL with stationary robots was first studied by Kurazume \emph{et al.}\cite{kurazume1994cooperative}\cite{Kurazume1996}. The author proposes a method called ¡±Cooperative Positioning System¡±(CPS). Within the CPS, a team of robots are divided into a stationary group, which acts as landmarks, and a mobile group. The roles of groups would reverse at the next time instant until the team reaches the destination. More related issues are discussed in \cite{kurazume1998study} and \cite{kurazume2000experimental}. A leader-assistive approach is presented in \cite{wanasinghe2014distributed} and \cite{wanasinghe2015distributed}, where a team of robots are divided into two groups, the performance-plus robots act as leader and the rest as child. The leaders have high-resolution estimators and help to localize the child robots. Such an approach is further extended in \cite{Nemsick2017}. The main limitation of these approaches is that too many constrains(such as the order of movement) are required on the robot motion. In our approach, each robot moves with its own path.

\section{System Setup and Problem Statement}
Consider a group of N robots navigating in 2D environment, each robot is equipped with proprioceptive and exteroceptive sensors. The state of individual robot at time $t$ is denoted as ${\bf{X}}_{\bf{t}} = {[{x_t},{y_t},{\theta _t}]^T}$, which contains the information of position and orientation. In this approach, we adopt the \emph{velocity motion model} with the assumption that a mobile robot can be controlled through a rotational and translational velocity denoted by ${v_t}$ and ${\omega _t}$ respectively. Thus, the control vector is denoted as ${{{\bf{u}}_{\bf{t}}}} = {[\begin{array}{*{20}{c}}{{v_t}}&{{\omega _t}}\end{array}]^T}$. It derives from the proprioceptive sensors such as encoder. On the basis of motion model, at time $t$, the state of robot can be presented as following equation:

\begin{small}
\begin{equation}
\begin{split}
{{\bf{X}}_{\bf{t}}} &= g\left( {{{\bf{u}}_{\bf{t}}},{{\bf{X}}_{{\bf{t - 1}}}}} \right) + N\left( {0,{{\bf{R}}_{\bf{t}}}} \right)\\
g\left( {{{\bf{u}}_{\bf{t}}},{{\bf{X}}_{{\bf{t - 1}}}}} \right) &= {{\bf{X}}_{{\bf{t - 1}}}} + \left( {\begin{array}{*{20}{c}}
{ - \frac{{{v_t}}}{{{\omega _t}}}\sin {\theta _{t - 1}} + \frac{{{v_t}}}{{{\omega _t}}}\sin ({\theta _{t - 1}} + {\omega _t}\Delta t)}\\
{\frac{{{v_t}}}{{{\omega _t}}}\cos {\theta _{t - 1}} - \frac{{{v_t}}}{{{\omega _t}}}\cos ({\theta _{t - 1}} + {\omega _t}\Delta t)}\\
{{\omega _t}\Delta t}
\end{array}} \right)\\
\end{split}
\label{eq1}
\end{equation}
\end{small}
where $\Delta t$ is the duration of a time step, $N\left( {0,{{\bf{R}}_{\bf{t}}}} \right)$ is the motion noise in an additive Gaussian with zero mean and covariance matrix ${\bf{R}}_{\bf{t}}$ .

Additionally, when one robot detects another by exteroceptive sensors at time $t$, the relative measurement ${{\bf{z}}_t} = {\left( {{{\begin{array}{*{20}{c}}{{r_t}}&\phi\end{array}}_t}} \right)^T}$, ${{r_t}}$ and ${{\phi _t}}$ are relative distance and bearing respectively between two robots. The measurement model for robot $i$ measuring robot $j$ at time $t$ is
\begin{equation}
\begin{split}
{\bf{z}}_t^{i,j} &= h\left( {{\bf{X}}_t^i,{\bf{X}}_t^j} \right) + N\left( {0,{{\bf{Q}}_t}} \right)\\
h\left( {{\bf{X}}_t^i,{\bf{X}}_t^j} \right) &= \left( {\begin{array}{*{20}{c}}
{\sqrt {{{(x_t^j - x_t^i{\rm{)}}}^{\rm{2}}}{\rm{ + }}{{(y_t^j - y_t^i{\rm{)}}}^{\rm{2}}}} }\\
{a\tan 2(y_t^j - y_t^i,x_t^j - x_t^i) - \theta _t^i}
\end{array}} \right)\\
\end{split}
\label{eq2}
\end{equation}
Here, ${{\bf{X}}_t^i}$ and ${{\bf{X}}_t^j}$ are the actual state of robot $i$ and robot $j$. $N\left( {0,{{\bf{Q}}_t}} \right)$ is the additive Gaussian noise with zero mean and covariance matrix ${{{\bf{Q}}_t}}$.

At each time step $t$, each robot is provided with an estimate ${{{\bf{\bar X}}}_{\bf{t}}}$ of its own state and the corresponding covariance matrix ${\bf{\bar \Sigma }}{}_{\bf{t}}$ which represents the uncertainty. The available data is odometry of each robot and all relative measurements between each two robots.
For centralized scheme, center robot stores and processes all data including the state and measurement of each robot. It can be denoted as:
\begin{equation}
\begin{split}
&{\bf{X}} = [{{\hat x}^1},{{\hat x}^2},{{\hat x}^3},{{\hat x}^4}....{{\hat x}^i}]\\
&{\bf{Z}} = [{z^{_{1,2}}},{z^{3,5}}......{z^{i,j}}]\\
&[{{\bar x}^1},{{\bar x}^2},{{\bar x}^3},{{\bar x}^4}....{{\bar x}^i}] = filter({\bf{X}},{\bf{Z}})\\
\end{split}
\end{equation}
For our decentralized scheme, each robot $i$ only stores its last state, odometry and measurement to estimate current state. Depending on the odometry and motion model, each robot $i$ has a predicted state:
\begin{equation}
{\bf{\bar X}}_t^i = g\left( {{\bf{u}}_t^i,{\bf{\bar X}}_{t - 1}^i} \right)
\end{equation}
When the robot $i$ have several relative measurements with $n$ robots. The objective of our approach is to find a proper estimator which meets the following equation:
\begin{equation}
\begin{split}
{\bf{\bar X}}_t^i &= \arg \min \sum\limits_{j = 1}^n {({\bf{z}}_t^{i,j} - \bar z_t^{i,j})} \\
&= \arg \min \sum\limits_{j = 1}^n {\left( {{\bf{z}}_t^{i,j} - h\left( {{\bf{\bar X}}_t^i,{\bf{\bar X}}_t^j} \right)} \right)}\\
\end{split}
\end{equation}

\section{Proposed Approach}

The purpose of our approach is to localize a group of mobile robots navigating in featureless environment depending on the relative measurement between two robots. We have the following assumptions: i) Each robot is equipped with proprioceptive and exteroceptive sensors to obtain data of odometry and relative measurement. ii) Each robot is marked with tags in order to be detected. iii) Robot is able to communicate with each other. In our localization algorithm, at any given time, firstly, one robot is chose randomly to remain stationary. Secondly, the moving robot which is closest to the stationary robot estimates its state by localization algorithm with available relative measurement. Sequentially, the rest of moving robots can utilize the measurement with former estimated robot. This process is usually broken into two steps, namely \emph{prediction} and \emph{correction}. In the following subsection, we would derive the equation of this process.

\subsection{Prediction}
Suppose that there are one stationary robot $j$ and several moving robots $i$ at time $t$. Robot predicts its state mainly by the odometry noted as ${{{\bf{u}}_t}}$. Thus, the predicted state of stationary robot $j$ is same as that at previous time step $t-1$:

\begin{equation}
\begin{split}
{\bf{\bar X}}_t^j &= {\bf{\bar X}}_{t - 1}^j\\
{\bf{\bar \Sigma }}_t^j &= {\bf{\bar \Sigma }}_{t - 1}^j\\
\end{split}
\end{equation}

Moving robots $i$ change its state during each time step. Therefore, depending on standard EKF, the predicted state and the corresponding covariance matrix can be derived by the following equation:

\begin{equation}
\begin{split}
{\bf{\bar X}}_t^i &= g\left( {{{\bf{u}}_t},{\bf{\bar X}}_{t - 1}^i} \right)\\
{\bf{\bar \Sigma }}_t^i &= {{\bf{G}}_t}{\bf{\bar \Sigma }}_{t - 1}^i{\bf{G}}_t^T + {{\bf{V}}_t}{{\bf{M}}_t}{\bf{V}}_t^T\\
\end{split}
\label{eq3}
\end{equation}

where, ${{\bf{\bar X}}_{t - 1}^i}$ and ${\bf{\bar \Sigma }}_{t - 1}^i$ is the estimated state and corresponding covariance matrix at time $t-1$. ${{\bf{G}}_t}{\bf{\bar \Sigma }}_{t - 1}^i{\bf{G}}_t^T$ represents the uncertainty of robot $i$ and ${{\bf{V}}_t}{{\bf{M}}_t}{\bf{V}}_t^T$ provides an approximate mapping between the motion noise in control space to the motion noise in state space. ${{\bf{M}}_t}$ is the covariance matrix of the noise in control space. And Jacobian ${{\bf{G}}_t}$ is the derivative of the function $g$ with respect to the state vector, ${{\bf{V}}_{\bf{t}}}$ is the derivative of the function $g$ with respect to control vector.

The estimate is calculated based on the odometry in the prediction step. However, this step is similar to DR that would have a cumulative error over a long distance. Hence, we need fuse the relative measurement to correct the estimate and reduce the uncertainty.

\subsection{Correction}
The order of the robots to be corrected  with the predicted state depends on the relative distance to the stationary robot $j$. We assume the closest one is robot $k$, its predicted state and corresponding variance matrix at time $t$ are ${\bf{\bar X}}_t^k$ and ${\bf{\bar \Sigma }}_t^k$. Hence, we have the prediction of relative measurement between $k$ and $j$:

\begin{equation}
{\bf{\bar z}}_t^{k,j} = h\left( {{\bf{\bar X}}_t^k,{\bf{\bar X}}_t^j} \right)
\label{eq6}
\end{equation}
We utilize the difference between the measurement ${\bf{z}}_t^{k,j}$ collected from exteroceptive sensors and the predicted measurement ${\bf{\bar z}}_t^{k,j}$ calculated in (\ref{eq6}) to correct the estimated state. There are three factors that would influence the accuracy of measurement, including the uncertainty of stationary robots, the uncertainty of moving robot and the Gaussian noise in exteroceptive sensors. Hence, we have:
\begin{equation}
{\bf{S}}_t^{k,j} = {\bf{H}}_t^k{\bf{\bar \Sigma }}_t^k{\left[ {{\bf{H}}_t^k} \right]^T} + {\bf{H}}_t^j{\bf{\bar \Sigma }}_t^j{\left[ {{\bf{H}}_t^j} \right]^T} + {{\bf{Q}}_t}
\end{equation}
Here, ${\bf{H}}_t^k$ is the Jacobian of $h$ with respect to the state vector of moving robot $k$. ${\bf{H}}_t^j$ is the Jacobian of $h$ with respect to the state vector of stationary robot. ${\bf{Q}}_t$ is the covariance matrix of measurement noise. This equation represents the noise mapping into the measurement space.

Then, depending on standard EKF, we can calculate the Kalman gain as following:
\begin{equation}
{\bf{K}}_t^{k,j} = {\bf{\bar \Sigma }}_t^k{\left[ {{\bf{H}}_t^k} \right]^T}{\left( {{\bf{S}}_t^k} \right)^{ - 1}}
\end{equation}

Fusing the previous formulas, the estimate ${{\bf{\bar X}}_t^k}$ and corresponding variance matrix ${\bf{\bar \Sigma }}_t^k$ of moving robot $k$ can be updated by the following equation:
\begin{equation}
\begin{split}
{\bf{\bar X}}_t^k &= {\bf{\bar X}}_t^k + {\bf{K}}_t^{k,j}\left( {{\bf{z}}_t^{k,j} - {\bf{\bar z}}_t^{k,j}} \right)\\
{\bf{\bar \Sigma }}_t^k &= \left( {{\bf{I}} - {\bf{K}}_t^{k,j}{\bf{H}}_t^k} \right){\bf{\bar \Sigma }}_t^k\\
\end{split}
\end{equation}
Here ${\bf{I}}$ is the identity matrix of the same dimensions as ${{\bf{K}}_t^k{\bf{H}}_t^k}$.

So far, we obtain the estimate state and corresponding covariance matrix of moving robot $k$ by fusing the odometry and relative measurement with stationary robot $j$ at time $t$. Essentially, we have the assumption that all probabilities of measurement between robots are independent that enable us to incrementally add the information from multiple robots into our filter\cite{thrun2005probabilistic}. Therefore, the rest of the robots can utilize the measurement of estimated moving robot like robot $k$ to estimate its state efficiently. For the remainder moving robot $m$ and $n$ estimated robot:

\begin{equation}
\begin{split}
{\bf{\bar X}}_t^m &= \sum\limits_{j = 1}^n {\left( {{\bf{K}}_t^{m,j}\left( {{\bf{z}}_t^{m,j} - {\bf{\bar z}}_t^{m,j}} \right)} \right)}\\
{\bf{\bar \Sigma }}_t^m &= \prod\limits_{j = 1}^n {\left( {{\bf{I}} - {\bf{K}}_t^{m,j}{\bf{H}}_t^m} \right)} \\
\end{split}
\end{equation}
Accordingly, the belief derive from the stationary robot propagates to all the robots that reduces the uncertainty of moving robot and improves the stability.

\subsection{Accuracy and Consistency}
The estimated state of robot is represented by the approximation and corresponding covariance matrix. To evaluate the accuracy of this estimator, we calculate the \emph{root mean square error}(RMSE) between two state vectors. The definition is as following:

\begin{equation}
RMSE: = \frac{1}{n}{\sum\limits_{i = 1}^n {\left\| {{{\bf{X}}_t} - {{{\bf{\bar X}}}_t}} \right\|} _2}
\end{equation}
${{{\bf{X}}_t}}$ and ${{{{\bf{\bar X}}}_t}}$ is the actual and estimated state. RMSE indicates how similar the actual and estimated state are over $n$ time steps.


\section{Simulation and Experiment}

\subsection{Simulation}
\begin{figure}
\centering
        \includegraphics[width=0.35\textwidth,clip]{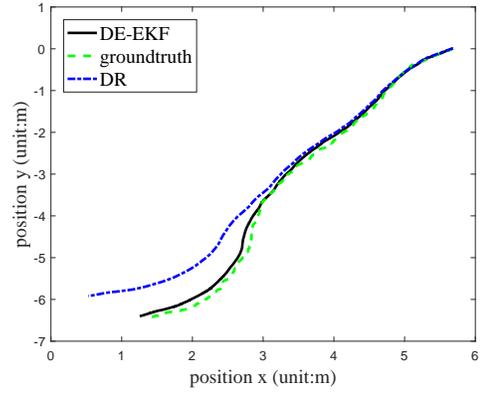}
    \caption{The trajectory of one robot calculated from DR and De-EKF and the groundtruth}
    \label{isimu_position}
\end{figure}

The performance of our approach was tested in simulation with $N=5$ robots navigating in an area of approximately 10 m$\times$10 m for 1000 time steps, each time step has a duration of 0.05 sec. In order to validate the performance, we have the following setting:
\begin{itemize}
\item The motion model and measurement model of all robots are identical which conforms to (\ref{eq1}) and (\ref{eq2}).
\item Each robot has individual translational velocity ${v_t} = 0.2$ m/s and rotational ${\omega _t} = 0.5*randn$ rad/s, the $randn$ is a normally distributed random number.
\item The initial pose error(m,m,rad)=[$0.01,0.01,0.01$], the initial actual position(m,m,rad)=[$5*randn,5*randn,2*randn$] and the measurement error(m,rad)=[$0.01,0.01$].
\end{itemize}
Under the above setting, the position of one robot calculated from our approach, decentralized EKF(De-EKF) and the dead-reckoning(DR) are shown in Fig. \ref{isimu_position}. We can see that, the DR and De-EKF have similar performance in the beginning. However, as the distance increases, estimator from DR have a cumulative error, while that from De-EKF accompanies the groundtruth constantly.

\begin{figure}[tpb]
\centering
    \begin{tabular}{cc}
        \includegraphics[width=0.23\textwidth]{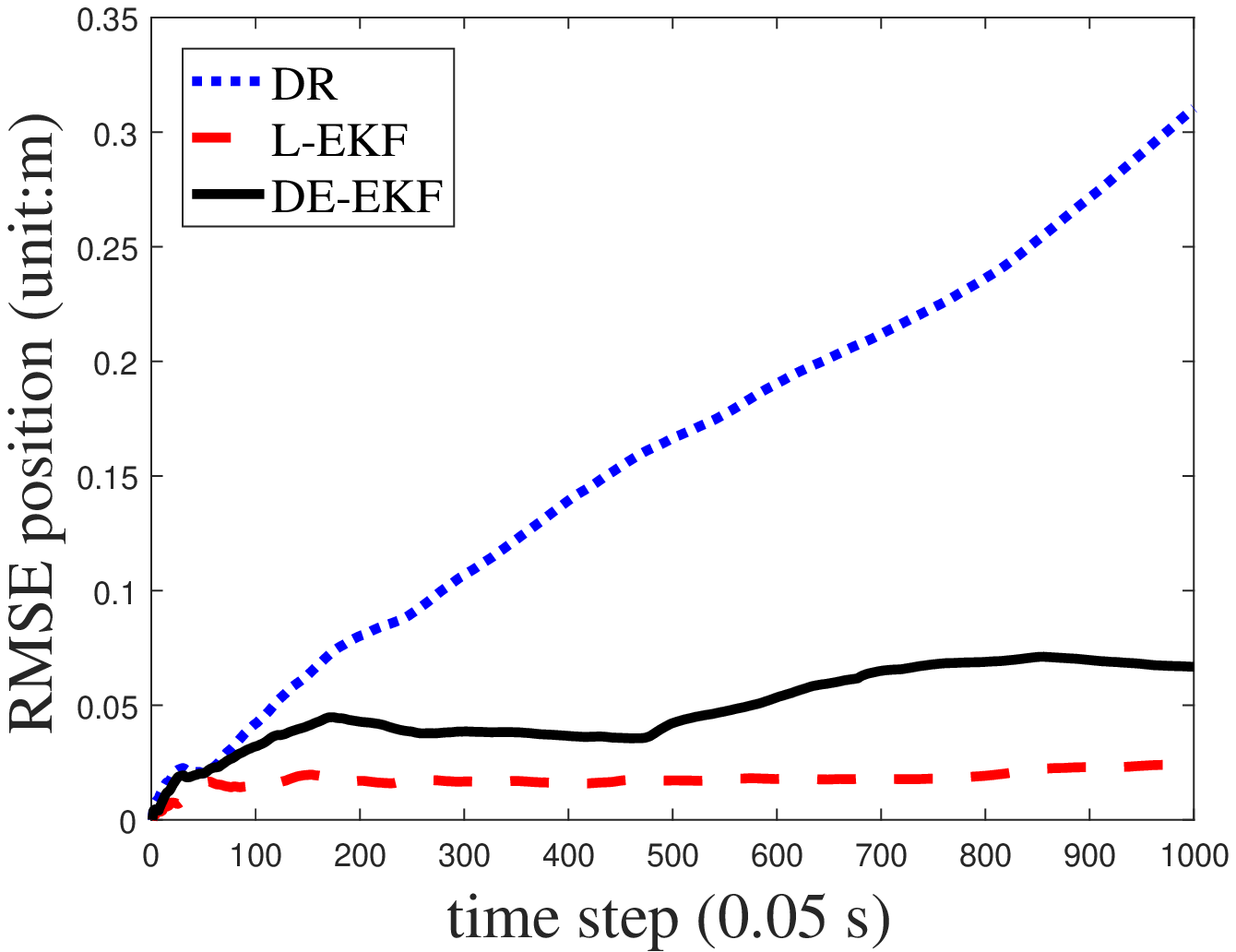}&
        \includegraphics[width=0.23\textwidth]{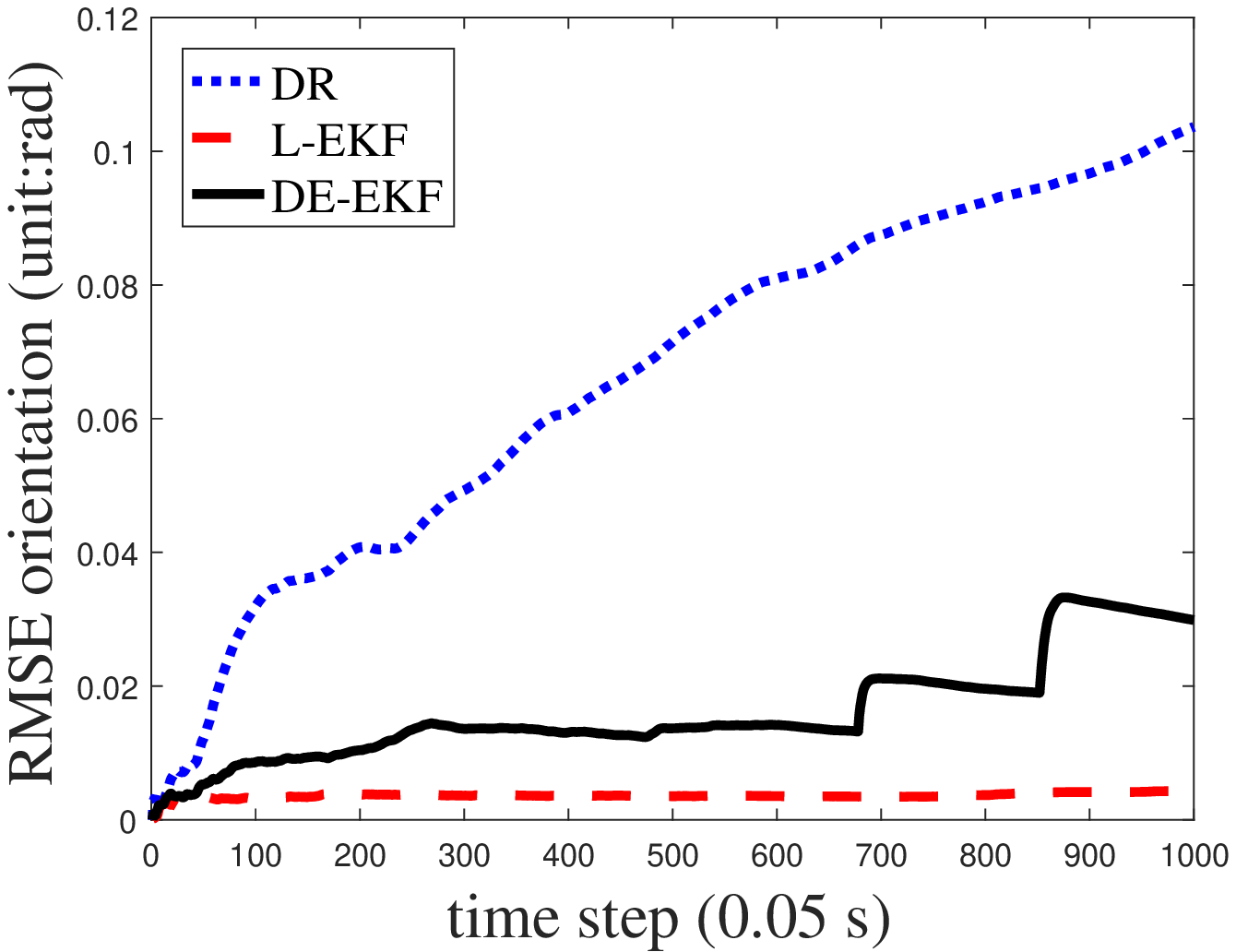}\\
        (a)&(b)\\
    \end{tabular}
    \caption{Comparison of DR, DE-EKF, L-EKF: (a) RMSE of robot position;
    (b) RMSE of robot orientation;}
    \label{compare}
\end{figure}

Then, we compare the performance of the following approaches: i) DR which estimates the state depending on odometry only. ii)L-EKF that one robot estimates its state by four known landmarks. iii)our approach(DE-EKF) that five robots estimates their states mainly utilize the relative measurement. We employ RMSE to indicate the accuracy of these approaches. The result of one robot is depicted in Fig. \ref{compare}. As expected, the DR has the worst performance, while the L-EKF has the best performance because the known landmark with absolute position can reduce the uncertainty of robot efficiently. Our approach is less accurate than the L-EKF, since the portable landmarks which are stationary robot or estimated robots have their own uncertainty. However, our approach do not need the known landmark. And it's capable for featureless environment. Additionally, the accuracy of our approach is close to that of L-EKF.

\begin{figure}
\centering
        \includegraphics[width=0.35\textwidth,clip]{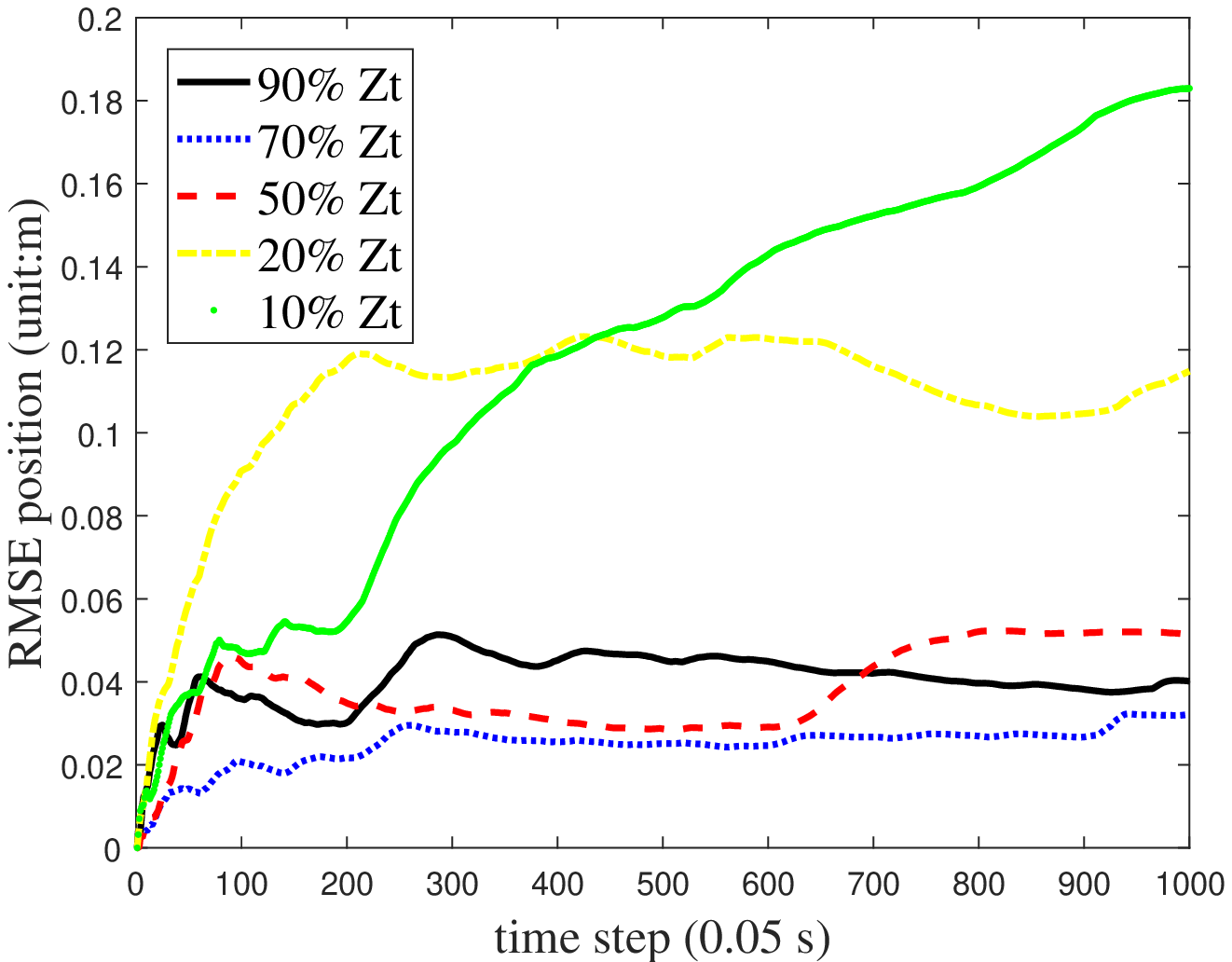}
    \caption{Comparison of the performance with insufficient measurements}
    \label{ED_Zt}
\end{figure}

In order to test the stability and robustness of our approach, we simulate the condition with insufficient relative measurement. The performance of our approach with 10\%-90\% measurement in each time step is compared in Fig. \ref{ED_Zt}. It depicts that with over 50\% available measurement, our approach has normal performance as usual. Even though there are 20\% available measurement, our approach still has a moderate performance that the RMSE in position is approximately 0.12m.

\subsection{Experiment}
\begin{figure}[htbp]
\centering
        \includegraphics[width=0.35\textwidth,clip]{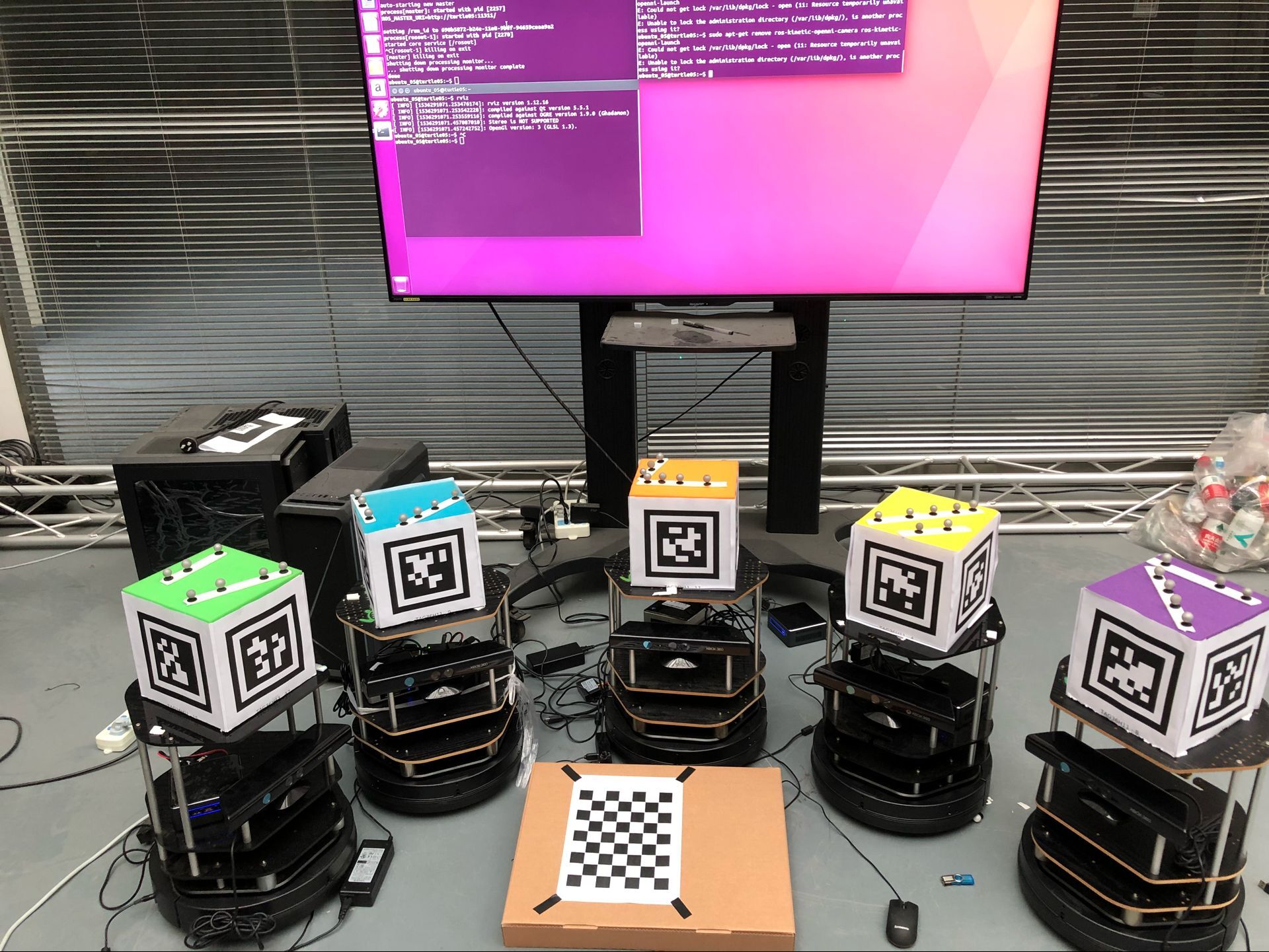}
    \caption{Structure of Turtlebot2}
    \label{turtlebot}
\end{figure}

An experiment with five modified Turtlebot2 is performed to validate the performance of our approach. The hardware and software used in experiment are specified in table \ref{Hardware} and \ref{Software}. Turtlebot2 is a low-cost personal wheeled robot kit with open-source software. The structure of modified Turtlebot2 is illustrated in Fig. \ref{turtlebot}. Each Turtlebot2 is equipped with encoder to acquire odometry. The exteroceptive sensors are structure light camera kinect and full HD camera located in front and back of this robot to acquire relative measurement. The actual information of robot position and rotation are provided by motion capture system(OptiTrack) with error less than 0.3mm and ${0.05^ \circ }$ respectively. Two features are applied on Turtlebot2. One is the rigid body marker, used to be detected by OptiTrack. Another is Apriltag, used to be detected by camera. The data of each robot is processed in mini PC which is a Inter NUC on our Turtlebot2.

\begin{figure}[H]
\centering
        \includegraphics[width=0.35\textwidth,clip]{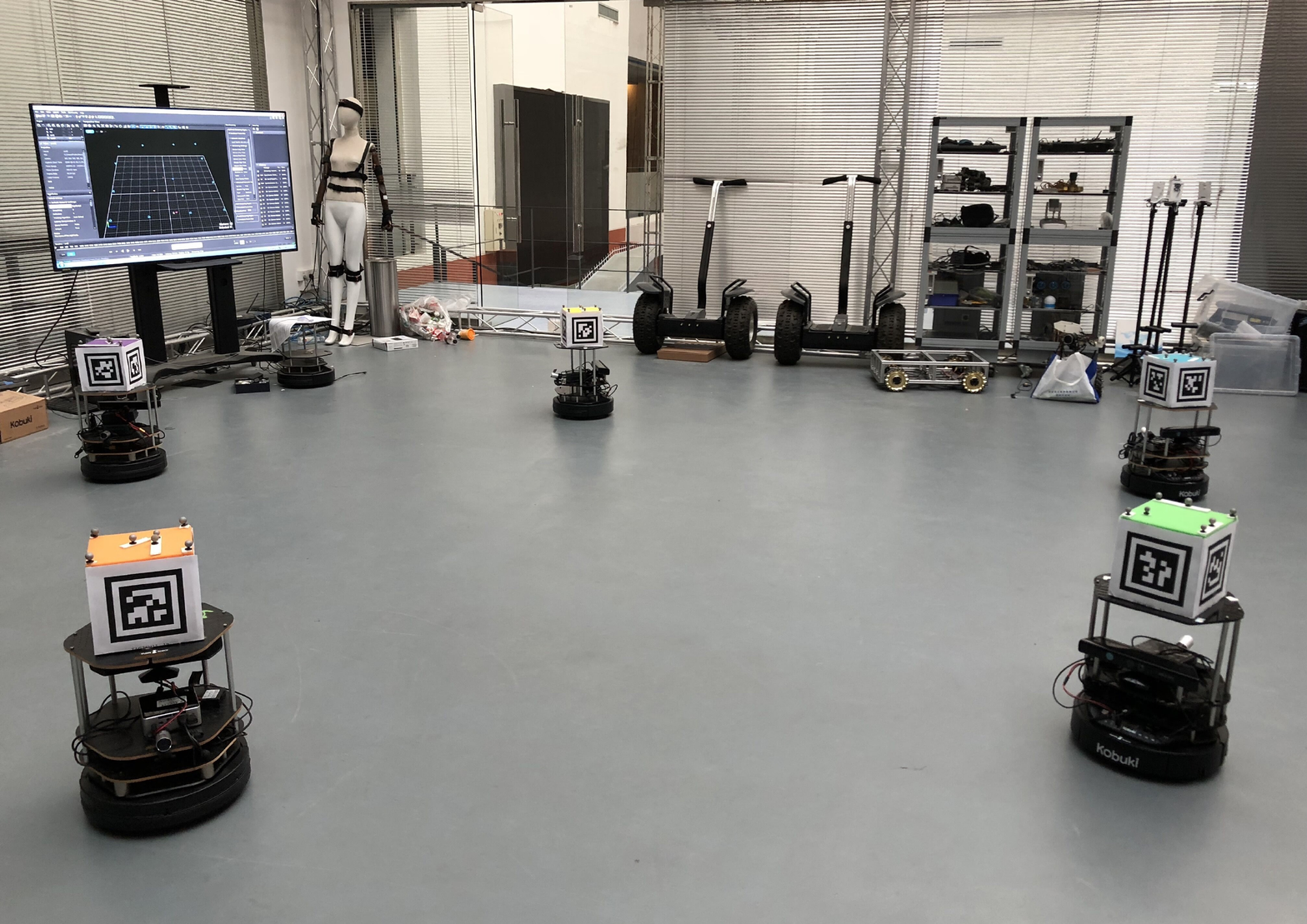}
    \caption{Experiment environment}
    \label{experiment}
\end{figure}

\begin{figure}[htbp]
\centering
        \includegraphics[width=0.35\textwidth,clip]{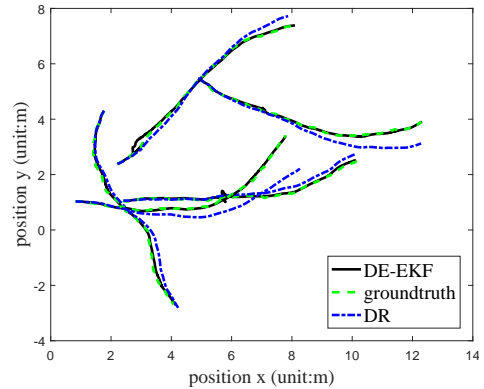}
    \caption{Trajectory of five robots}
    \label{position}
\end{figure}

\begin{figure}[htbp]
\centering
    \begin{tabular}{cc}
        \includegraphics[width=0.23\textwidth]{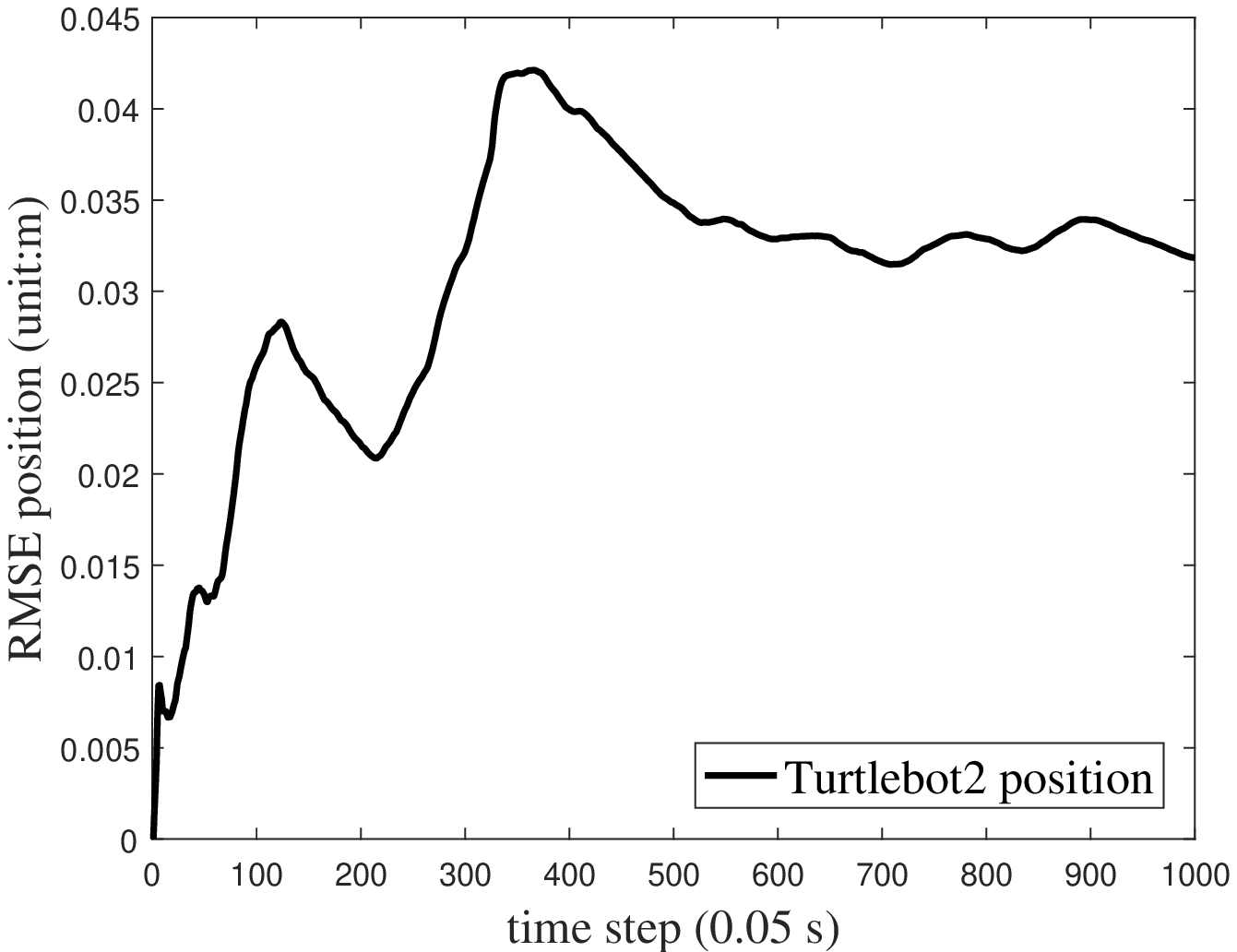}&
        \includegraphics[width=0.23\textwidth]{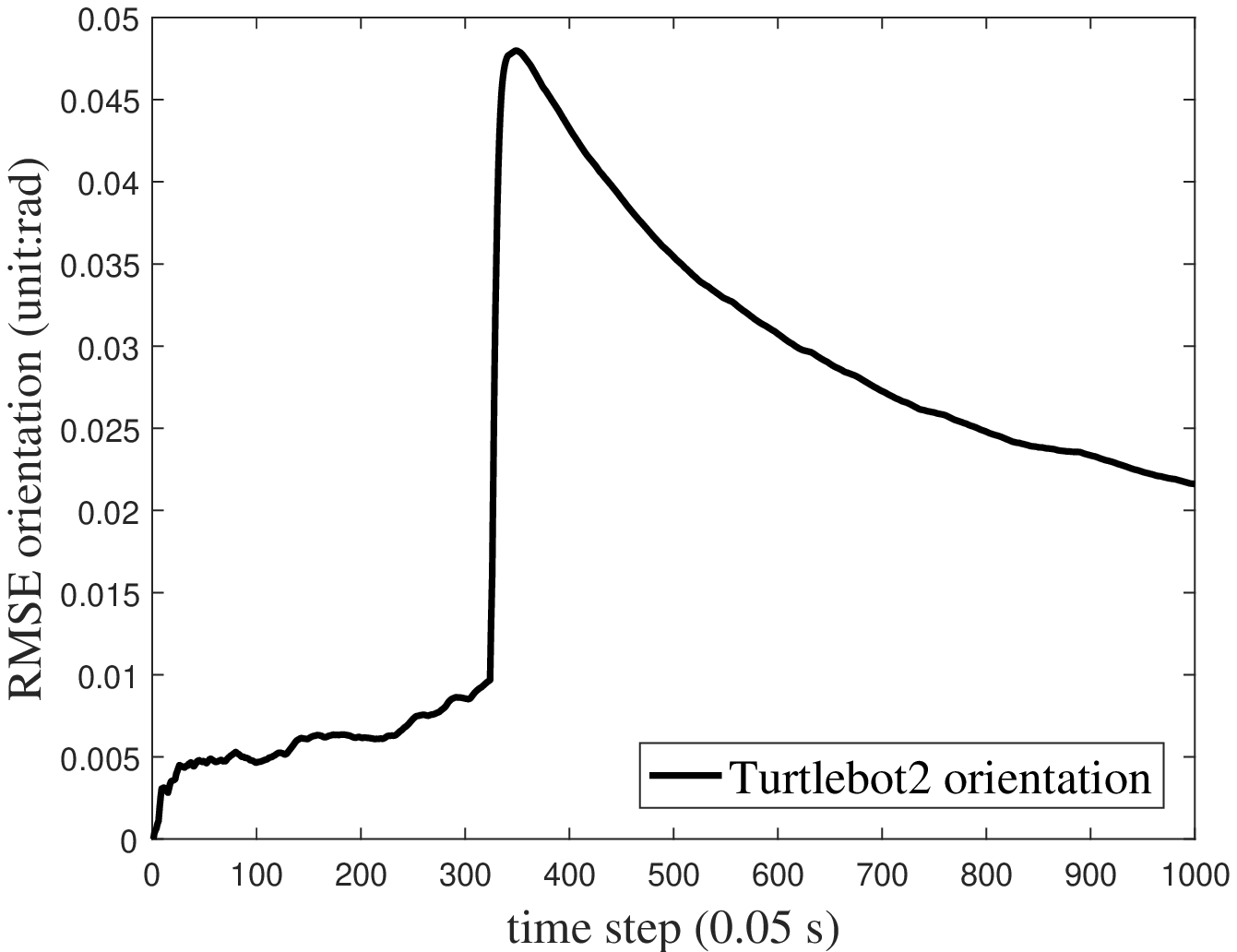}\\
        (a)&(b)\\
    \end{tabular}
    \caption{experiment result(a) RMSE of robot position;
    (b) RMSE of robot orientation;}
    \label{RMSE}
\end{figure}

The experiment environment is shown in Fig. \ref{experiment}. Five Turtlebot2 navigate in this environment with the motion strategy as we mentioned previously. The experiment result is shown in Fig. \ref{position} and Fig. \ref{RMSE}. We can see the state of five robots calculated by our approach approximates to the groundtruth. The RMSE of position and orientation is less than 0.04m and 0.04 rad respectively.

\begin{table}[htbp]
\scriptsize
\renewcommand{\arraystretch}{1.3}
\setlength{\abovecaptionskip}{0pt}%
\setlength{\belowcaptionskip}{0pt}
\caption{Hardware Used in Experiment}
\label{Hardware}
\begin{center}
\begin{tabular}{|c||c||c|}
\hline
Hardware & Type & Description\\
\hline
Turtlebot2 base & YMR-K01-W1 &Provide Odometry\\
\hline
Full HD Camera & Microsoft LifeCam 1425 & Relative pose detection\\
\hline
Kinect & Microsoft Kinect V1 & Relative pose detection\\
\hline
rigid body marker & - & Unique feature\\
\hline
Apriltag & Tag36h11 & Unique feature\\
\hline
Intel NUC Kit & J57645-309 & Data processing\\
\hline
Optitrack Camera & Prime 13 & Provide ground truth\\
\hline
\end{tabular}
\end{center}
\end{table}

\begin{table}[htbp]
\scriptsize
\renewcommand{\arraystretch}{1.3}
\setlength{\abovecaptionskip}{0pt}%
\setlength{\belowcaptionskip}{0pt}
\caption{Software Used in Experiment}
\label{Software}
\begin{center}
\begin{tabular}{|c||c||c|}
\hline
Software & Version & Description\\
\hline
Optitrack Motive & 2.0 & System driver\\
\hline
usb cam & 0.3.6 & ROS camera driver\\
\hline
Openni camera driver & 1.11.1 & ROS Kinect driver\\
\hline
Apriltags2 ROS & 0.9.8 & Provide relative pose\\
\hline
ROS & Kinetic & Robot operating system\\
\hline
Mocap Optitrack & 2.8.3 & Optitrack driver\\
\hline
\end{tabular}
\end{center}
\end{table}

\section{Conclusion and Future Work}
In this paper, a decentralized approach with improved EKF for cooperative localization is proposed. Compared with other multi-robot localization methods, the proposed approach is advantageous in high accuracy, scalability and robustness of state estimation, as well as low hardware cost and computational complexity. This approach provides a solution for localizing a group of robots in featureless environments. The experiment results show that the proposed approach can yield similar localization accuracy as the landmark-based method does, without using any pre-known landmarks. For five robots, the RMSE of position and orientation is less than 0.04m and 0.04 rad respectively.  When the measurements are insufficient, the proposed approach still can yield a moderate localization performance. However, the system performance would decline when the robots move at high speeds. Future work will be focused on addressing this problem through using high sampling rates and faster estimation algorithms.


\bibliographystyle{IEEEtran}
\bibliography{reference/Thesis}
\end{document}